%% file: Chart-Pattern-detection-using-Machine-Learning-algorithms.tex
\documentclass[twocolumn]{article}
\usepackage[top=2cm, bottom=2.5cm, left=2cm, right=2cm]{geometry}
\usepackage{times,textcomp}
\usepackage[T1]{fontenc}
\usepackage[utf8]{inputenc}
\usepackage{amssymb,amsmath}
\usepackage{enumitem}
\usepackage{eurosym}

\usepackage{titlesec}

\usepackage{titling}
\newcommand{\subtitle}[1]{
	\posttitle{
		\par\end{center}
	\begin{center}\large#1\end{center}
	\vskip0.5em}
}

\usepackage{titlesec} 
\titleformat{\section}{\large\bfseries}{\thesection}{1em}{\MakeUppercase} 
\titleformat{\subsection}{\large\bfseries}{\thesubsection}{1em}{}
\titlespacing*{\section}{0pt}{\parskip}{0pt}
\titlespacing*{\subsection}{0pt}{\parskip}{0pt}

\usepackage{longtable,booktabs}
\usepackage{graphicx,grffile}
\usepackage{fancyhdr}

\title{\huge\bfseries Stock Chart Pattern recognition with Deep Learning}

\usepackage{authblk}

\author{Marc Velay}
\author{Fabrice Daniel}

\affil{\small Artificial Intelligence Department of Lusis, Paris, France\\fabrice.daniel@lusis.fr\\http://www.lusis.fr}

\date{June 2018}
\begin{document}

\maketitle

\begin{abstract}
This study evaluates the performances of CNN and LSTM for recognizing common charts patterns in a stock historical data. It presents two common patterns, the method used to build the training set, the neural networks architectures and the accuracies obtained.
\end{abstract}


\input{chapters/introduction}
\input{chapters/target-pattern}
\input{chapters/approaches}
\input{chapters/results}
\input{chapters/problems}
\input{chapters/conclusion}

\input{chapters/reference}
\end{document}

%% file: chapters/introduction.tex
\section{Introduction}

Patterns are recurring sequences found in OHLC\footnote{Open High Low Close} candlestick charts which traders have historically used as buy and sell signals. Several studies, notably by Bulkowski\footnote{http://thepatternsite.com/}, have found some correlation between patterns and future trends, although to a limited extent. The correlations were found to be between 50 and 60\%, 50\% being no better than random. Many traders are using chart patterns, sometimes combined with other techniques, to take their trading decisions, in a field known as Technical Analysis.

Our goal is to automate the detection of these patterns and to evaluate how a Deep Learning based recognizer behaves compared to hard-coded one. Automation would simplify the process of finding sequences which vary in scale and length. It would also help provide valuable information for stock market price prediction as these signals do offer small correlation with prices\cite{TA}\cite{TS-detection}. Alone, the patterns are not enough to predict trends, according to other studies, but may yield different results when coupled with other indicators. 

Several types of detection algorithms exist ranging from pragmatic to machine learning. The solutions vary in efficiency, re-usability and speed, in theory.

The first solution is an hard-coded algorithm. It is fast to implement and gives quick results, but requires a human to find parameters for any given pattern. Extending the number of patterns recognized implies having a human examining candlestick charts in order to deduce signal characteristics before implementing the detection using conditions specific to that pattern. Adding different parameters to modulate ratios allows us to tweak the patterns' characteristics. This technique does not have any generalization potential. If the pattern is slightly outside of the defined bounds, it will not be detected, even if a human would have classified it otherwise.

Another solution is DTW\footnote{Dynamic Time Warping} which consists in computing the distance between two time series. DTW allows us to recognize a pattern that could vary in size and length. To use this algorithm, we must use reference time series, which have to be selected by a human. The references must generalize well when compared with signals similar to the pattern in order to capture the whole range.

The solution we propose to study is based on Deep Learning. There exists several ways to detect patterns in time series using neural networks. First, we have evaluated a 2D CNN\footnote{Convolutional Neural Networks} applied to charts images. This is the state of the art in many image related applications, such as facial recognition. Then two other networks, well suited to time series application, were compared; a 1D CNN and a LSTM\footnote{Long Short-Term Memory}, a type of recurrent neural network widely used for state of the art NLP\footnote{Natural Language Processing}. In addition to their accuracy, our purpose is also to measure the generalization potential of these Deep Learning models compared to hard-coded solution.

The detection of chart patterns, in order to build a strategy or notify users, is not a simple problem. In either case, false positives have a very negative effect, either wasting a user's time or ruining a trading strategy. An hard-coded algorithm, dependant of manually selected parameters, is able to detect most obvious patterns. We have selected bounds which are strict enough to reduce the type I error, false positives, to nearly 0. In exchange we have a higher type II error rate, false negatives. This type of error imply missed opportunities and is therefore less important than acting at wrong times. 
We want to analyze the effect of Deep Neural networks on the detection of patterns. In theory, this solution should be able to keep a low type I error rate, while also reducing the type II error rate. This effect would be reached by learning a more general shape of the pattern. We plan to measure this effect through the recall and generalization rates. The recall rate is defined as the ratio of actual patterns detected compared to the total amount of patterns detected, the true positive rate. It tracks how close we are to the manually parameterized solution. Generalization is defined as the gap between the amount of actual patterns detected and the expected amount of detected patterns, defined by the hard-coded algorithm's labels. Generalization has to be measured manually, as the labels are not always accurate, by analyzing the false negative. To efficiently determine the generalization rate, the model needs to be able to detect most patterns the hard-coded algorithm labeled as true.
The gain in accuracy provided by a higher generalization must be greater than the loss of accuracy provided by a recall near 100\%. The DNN must, in fact, be able to detect at least as many patterns as the manually parameterized solution, and should detect more patterns by generalizing the shape.

%% file: chapters/target-pattern.tex
\section{Target pattern}

\begin{figure*}
\centering
\includegraphics[scale=0.4]{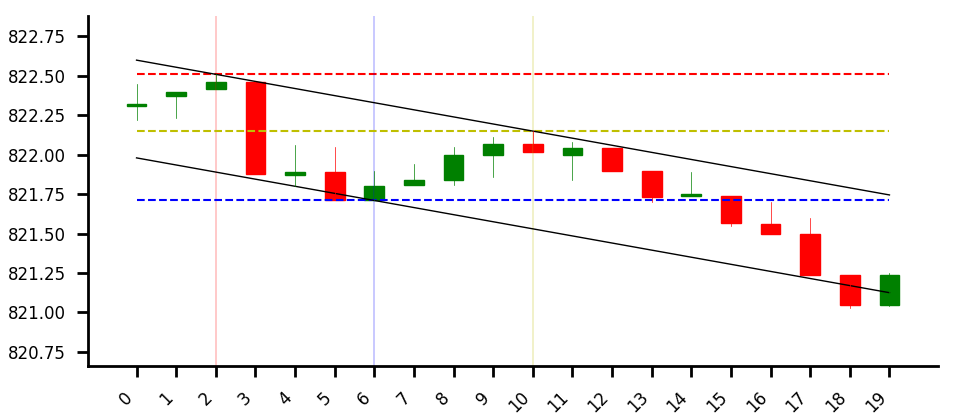}
\caption{{\it Bearish flag pattern\/}}
\label{bearish-flag}
\end{figure*}

The first pattern we studied was a bearish flag, where we can observe an initial drop from an extrema in the price called a flagpole, followed by a high volatility, slow increase in price called a flag and then followed by another flagpole of similar length to the first one. This is illustrated in Figure \ref{bearish-flag}. This pattern seems to be associated with a downwards trend and has a rather distinctive look, which would make detection easier.

\begin{figure}
\centering
\includegraphics[scale=0.5]{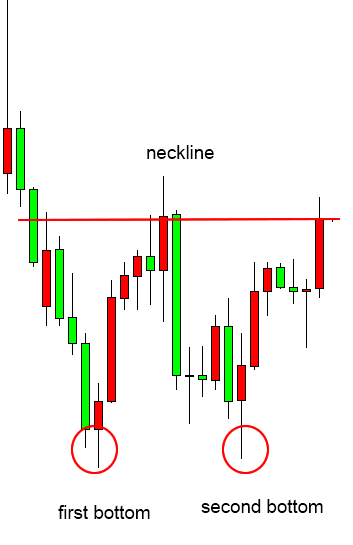}
\caption{{\it Double bottom pattern\/}}
\label{double-bottom}
\end{figure}

The next patterns we analyzed are "double top" and "double bottom". They present two extremas separated by an opposite local extrema. For these patterns the peaks have equal or close values and the reversal between them is seen as a pullback value, which has to be crossed again in order to complete the pattern. This pattern is illustrated in Figure \ref{double-bottom}. The patterns have been extensively analyzed \cite{DoublePatterns} and offer a profit:risk ratio of 1 with commonly accepted trading strategies having been developed around them. The strategies generally involve using the pullback value as a stop loss and using twice its distance to the extremas as the price target which should be met before selling. The real profit from those strategies has not been disclosed in resources found during our search. It would therefore be interesting to run a trading simulation using historical data in order to confirm its efficiency.

%% file: chapters/approaches.tex
\section{Possible approaches}

The research process consists in two steps. We first implement a hard-coded recognizer having some leeway in its detection model. We use it to build our training set. The second step consists on training several Deep Learning models on this training set. We then evaluate their generalization capabilities by observing if the Deep Learning approaches allows to detect not only the patterns of the training set but also some patterns that cannot be detected by the hard-coded model, due to scaling or position variations in the samples.

The hard-coded recognizer uses margins to detect the pattern with varying ratios between the segments. We can vary the margins in order to detect more or less patterns over the range. We ran this detector as a moving window over a specified time frame with a step value of 1 minute. The time frame used meant giving more or less historic information for predictions. These values ranged from 15 minutes to 3 hours. Each sample is a time frame of the OHLC values with a class associated to weather or not the pattern was present during the time frame. We based this analysis on data from Alphabet C stock from January 2017 to march 2018, with 1 minute intra-day data.

After building the training set, we starts training the CNN then the LSTM.

\begin{figure*}
\centering
\includegraphics[scale=0.4]{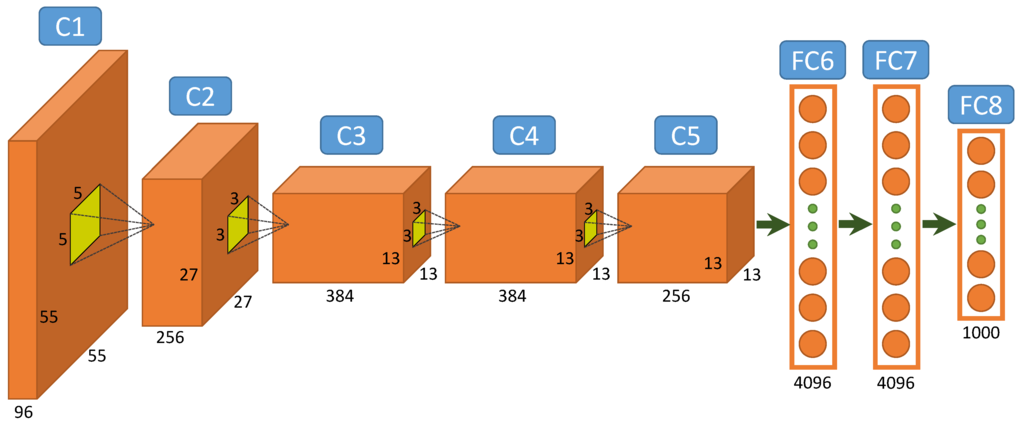}
\caption{{\it Convolutional Neural Network architecture\/}}
\label{cnn-archi}
\end{figure*}

A Convolutional Neural Network is a feedforward network which reduces the input's size by using convolutions. There has been some success with this technique already for this type of problem\cite{ForexTrendCNN}. We will implement both a 1D and 2D CNN. The 1D network expects a sequence with several channels, here the OHLC values. The 2D network, which is the state of the art for image processing, expects a matrix. This implies that several functionality have been encoded into its architecture in order to better extract information from 2D images \cite{CNN-lesson}\cite{CNN}. Our own choice of architecture is that of AlexNet, the architecture of the network is illustrated in Figure \ref{cnn-archi}. This CNN is widely perceived as being one of the most efficient for analyzing images. It relies on several connected convolution layers followed by fully connected layers. Since the network expects structures similar to images, we will use vignettes with information usually formatted for human consumption, such as candlestick and line graphs.

\begin{figure*}
\centering
\includegraphics[scale=0.4]{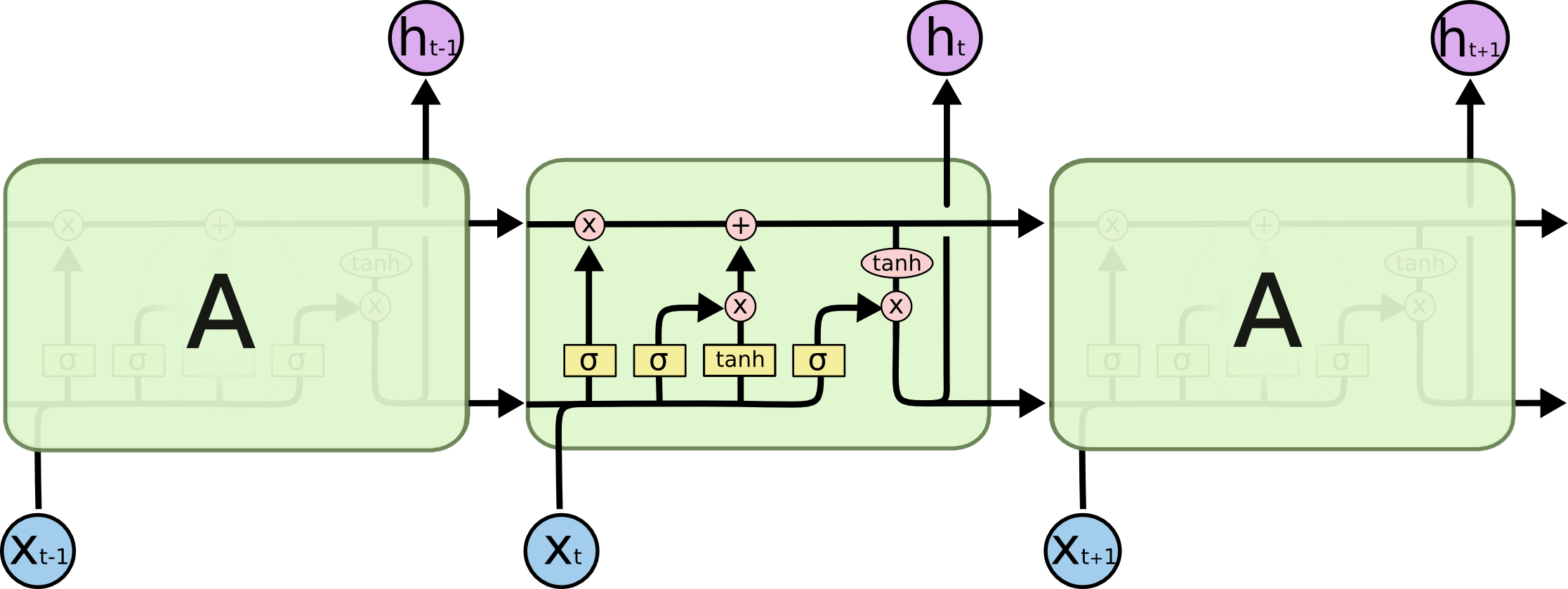}
\caption{{\it Long Short-Term Memory network architecture\/}}
\label{lstm-archi}
\end{figure*}

A Long Short-Term Memory recurrent network relies on past states and outputs to make predictions, we illustrate its architecture in Figure \ref{lstm-archi}. The main goal of a LSTM is to keep information that might be useful later in memory. Therefore it learns the relation through time between elements. The core nodes are made up of a state which is updated via several gates such as the update and forget gates and is used to compute the output for each future timestep. The state makes remembering links between elements easier, such as in patterns over time\cite{LSTM}. 

\begin{figure}
\centering
\includegraphics[scale=1.2]{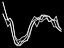}
\caption{{\it OHLC curves used to train the CNN\/}}
\label{ohlc-lines}
\end{figure}

Using the dataset built by the hard-coded recognizer, we can either transform it into images to train the 2D CNN or use it "as is" to train the LSTM or the 1D CNN. The transformation into images is done by plotting the values during the time frame, such as in Figure \ref{ohlc-lines}.

The vignettes are very small in size and have been adjusted above to be clearer for humans. Most of the information is clearly available to the network, even at this scale. A larger size would only mean more memory and time used up for no greater results (the variation in accuracy per class was too small to notice over several training attempts of over 50 epochs).

\begin{figure}
\centering
\includegraphics[scale=1.2]{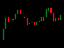}
\caption{{\it Candlestick graph used to train the CNN\/}}
\label{ohlc-candlestick}
\end{figure}

We eventually also attempted to train the CNN with traditional candlestick charts, such as Figure \ref{ohlc-candlestick}, which are used by humans to detect such patterns. It turns out that on average, using candlestick charts was ~3\% more efficient than the line charts.

For both the CNN and LSTM, we looked at the correlation of each variable (OHLCV) with the detection of the pattern. This is done to avoid giving too much or too little information to the model, which would lead to a reduced efficiency. We found that the closing value was most strongly correlated.

%% file: chapters/results.tex
\section{Results}

The hard-coded recognizer detected roughly 3000 occurrences of the bearish flag over the lapse of a year, with a window period of 30 minutes, which was the time-frame with the best results. The detected patterns were the ones which met the rough proportions determined by looking at well documented occurrences found on websites focused on Technical Analysis. The proportions being rather large, there might be a certain amount of false positives, therefore we chose to restrict the proportions. We have chosen the current parameters for detection in order to only detect actual occurrences, thus minimizing the initial false positives. We then used those detections as a training dataset for the deep learning algorithms. During training, we down-sampled the amount of examples without a pattern in order to have a 50\% distribution of both negative and positive occurrences. The samples were mostly evenly distributed over the total population, in order to avoid overlapping frames. The validation and training sets are sub-sets of our examples, picked at random.

This system can be used to launch alerts to users or even place buy and sell orders. False positives mean either disturbing a user for nothing or potentially losing money. Either of these outcomes must be avoided. Type II errors, or false negatives, only mean losing out on opportunities. Therefore, we must focus on reducing Type I error, yet keep type II error as low as possible.

When we analyzed the results of the LSTM model for the detection of patterns, we relied on a sliding window of fixed size and normalized OHLCV values. After optimizing the hyper-parameters, we found that a time-frame of 30 minutes was best suited for detecting patterns. We chose this length of time as the basis for all other algorithms. Too small a time-frame and the patterns were not complete, too large and the neural network could not extrapolate information due to a large amount of noise around the pattern. Using a simple LSTM, with a single layer of 10 units, we had a recall of 96.8\%, where 50\% is the accuracy of randomly pickling positive or negative.
Our goal is to have the highest recall so the lowest amount of false negatives. When testing on 1536 samples, from the validation set, the model only predicted 2 false positives. This is a 0.13\% false positive rate. The false negative rate was 1.4\%. We can minimize these values by filtering the patterns by hand, studying what makes the neural network predict one outcome over the other. But this is an expensive and time-consuming process. Studying a sample of the predictions, especially the false positives and false negatives, we found that the a part of the false positives were in fact true positives, which is most likely due to the hard-coded algorithm's parameters. In the false negatives, we mostly found actual miss-classification, due to the remaining error rate of the LSTM model. Therefore, we can conclude that the LSTM model is able to detect several occurrences of the pattern which the hard-coded algorithm had not itself detected. But, there is still a rather high error rate, which could be reduced by collecting more examples of the pattern and continuing to optimize the hyper-parameters of the network.

We used several tactics to train the 2D-CNN, such as finding the best hyper parameters using grid-search for different types of training data. This was a complicated process as there is a very narrow margin between over-fitting and under-fitting in our case, when and if the CNN finds a relation between the output and input. Using line charts, the best recall rate we found is 71\%. We obtained those results when we used a single line chart of the High value, similar to Figure \ref{ohlc-lines}. Using OHLC candlestick such as Figure \ref{ohlc-candlestick}, the best recall rate we found was 73\%. These cross-categorical scores are due to several factors, such as poor input data comprised of large sparse matrices. To improve the results, we should find an alternate data representation scheme which does not rely on large sparse matrices, such as using 1D CNN. The model had too much difficulty learning the relationship between input and output. Since the accuracy and recall rate are so low, with large amounts of false positives and false negatives, we can not compare the generalization potential of this model with that of an hard-coded algorithm. This deep learning model detects less occurrences of patterns than an hard-coded algorithm.

The 1D CNN used an identical input as the LSTM. A sequence of several variables. This proved to be the worse model out of the three. After tweaking the hyper-parameters and using identical data as the previous model, we only reached a 64\% recall rate. Similarly to the 2D CNN model, this recall rate is too low to draw correct conclusions about its generalization potential. We must conclude that CNN models do not provide better detection rates than hard-coded algorithms. 

\begin{table}[ht]
	\centering
	\begin{tabular}{lcr}
		\hline
		{\textbf{Algorithm}} & {\textbf{Recall}} & {\textbf{Generalization}} \\ \hline
		\textbf{LSTM}     & \textbf{0.97} & 0.3\% \\
		2D CNN            & 0.73          & --\\
		1D CNN            & 0.64          & --\\ \hline
	\end{tabular}
	\caption{Per algorithm recall rate} 
	\label{tab:results}
\end{table}

%% file: chapters/problems.tex
\section{Problems encountered}

The main issue we encountered was the quality of the data used. There are no pre-existing datasets with labeled patterns that could be found and we had to create them. The best way to do this regarding speed and quality was by building an hard-coded detector. This guaranteed a level of quality identical for every iteration that manual sorting can not provide. Yet the selection was done using handpicked parameters which imply that not all variations of the pattern were captured. If the machine learning algorithms generalize the shape of the pattern then it may lead to a misleading lower accuracy as it detects real patterns which the initial algorithm used to build the training set had not detected, so making these patterns wrongly misclassified. That's why false negative and false positive must be fixed manually before to compute the confusion matrix.

The type of data we give the 2D CNN is also important: it is comprised of sparse binary matrices with few values set to one. A majority of the pixels are set to 0 and the line in line graphs rarely occur in the same part of the graph. This makes it very hard for a CNN to figure out what the shape looks like as most samples will have few non-null values. Therefore the CNN has a trouble finding a correlation between the input and output. Since there is no evident segmentation, the optimizer can not find an activation matrix corresponding to a specific area in each convolution layer. The result is a very thin area between over-fitting and under-fitting where the model actually learns what the pattern is, which is explored by modifying the hyper-parameters. This problem was solved by using grid-search, but the time required to find decent parameters is very large.

We attempted to test the model on unrelated data, stock from another company. This gave us very poor results, which leads us to believe a trained model from one type of dataset will not generalize well to other datasets, even though we are looking for the same pattern and the data has been normalized. This problem has not been solved at the moment, but we believe it could be mitigated in future steps by training the models using several different datasets, such as equity from several companies.

%% file: chapters/conclusion.tex
\section{Conclusion}

We have achieved a study of detecting a single pattern in time series and comparing the generalization potential of several models compared to an hard-coded algorithm. We have found that the LSTM model achieved the best detection rates. The 1D and 2D CNN models were not able to reach high accuracy levels and could therefore not be compared to the hard-coded algorithm to evaluate their generalization potential. We manually sifted through the false negatives and positives predicted by the LSTM model. We found that it was able to somewhat generalize, but that it still had misclassified several patterns due to a 1.2\% error rate. There are several steps that could be taken in order to improve the results we have found so far.

The most important next step is to add more patterns to be recognized, such as in multi-task learning, which has proven benefits compared to single objective learning\cite{MTL}. This could be done by adding more classes or finding a suitable dataset containing the other patterns. This has not been attempted yet due to a difficulty behind the concept of “adding other patterns”. There would exist occurrences where several patterns would be present in the same time-frame.

If we manage to capture every pattern in a given dataset more efficiently than the current technique, we could also consider different techniques that we have not applied here, such as Dynamic Time Warping or comparing acoustic fingerprinting. The previous techniques rely on time-series and DTW has long been the state of the art in finding patterns. They would provide a benchmark for the results we have found using both the CNN and LSTM.

Encoders are popular for their ability to remove noise from data and dimension reduction tasks. One way to improve the CNN’s detection rate would be to use encoders in order to reduce the sparse matrices to something from which the model could extrapolate information from. They could also be used to “clean” the time-series we feed into the LSTM so that it could learn the shapes of the patterns instead of also learning the noise due to the volatility of the prices we use.

Once this study has been improved to get better results and generalizes to other datasets, we could consider adding it to a multimodal model for trend forecasting. There is a small correlation between the different patterns and the rise and falls of prices. Therefore adding it to a broader system could provide a little information that will prove helpful.